\documentclass[11pt,twocolumn]{article}
\usepackage{pslatex, graphicx, amssymb, amsmath}

\usepackage{multirow}
\usepackage[table]{xcolor}
\usepackage{booktabs}

\def\abstract{
\typeout{Abstract}
 {\bf Abstract} 
}

\begin{document}
\title{A positive feedback method based on F-measure value for Salient Object Detection}

\author{
	Ailing Pan \textsuperscript{1}\and Chao Dai \textsuperscript{1} 
	\and Chen Pan \textsuperscript{1} \thanks{Corresponding author. Email: pc916@cjlu.edu.cn} 
	\and Dongping Zhang \textsuperscript{1} \thanks{This work was supported by Key Research and Development Projects in Zhejiang Province of China (NO.2021C03192, 2023C01032).}
	\and Yunchao Xu \textsuperscript{1} 
	\\
	\small{\textsuperscript{1}Department of Computer Engineering, China JiLiang University, Hangzhou, Zhejiang, 310018, China}
}

\maketitle
\begin{abstract}
The majority of current salient object detection (SOD) models are focused on designing a series of decoders based on fully convolutional networks (FCNs) or Transformer architectures and integrating them in a skillful manner. These models have achieved remarkable high performance and made significant contributions to the development of SOD. Their primary research objective is to develop novel algorithms that can outperform state-of-the-art models, a task that is extremely difficult and time-consuming. In contrast, this paper proposes a positive feedback method based on F-measure value for SOD, aiming to improve the accuracy of saliency prediction using existing methods. Specifically, our proposed method takes an image to be detected and inputs it into several existing models to obtain their respective prediction maps. These prediction maps are then fed into our positive feedback method to generate the final prediction result, without the need for careful decoder design or model training. Through applying the positive feedback method for decision fusion of multi-model perception results. Moreover, our method is adaptive and can be implemented based on existing models without any restrictions. Experimental results on five publicly available datasets show that our proposed positive feedback method outperforms the latest 12 methods in five evaluation metrics for saliency map prediction. Additionally, we conducted a robustness experiment, which shows that when at least one good prediction result exists in the selected existing model, our proposed approach can ensure that the prediction result is not worse. Our approach achieves a prediction speed of 20 frames per second (FPS) when evaluated on a low configuration host and after removing the prediction time overhead of inserted models. These results highlight the effectiveness, efficiency, and robustness of our proposed approach for salient object detection. Code and Saliency maps will be available. 
\end{abstract}

\section{Introduction}
Salient object detection (SOD) aims to mimic the human visual perception system to capture the most prominent regions in given images or videos. It can serve as a pre-processing step for other realated computer vision tasks, including object tracking\cite{bib1}, action recognition\cite{bib2}, video segmentation\cite{bib3}, and image captioning\cite{bib4}.

Existing SOD models can be mainly divided into traditional algorithm-based and deep learning-based approaches. Traditional SOD methods rely on handcrafted features and use these features to predict saliency maps in a bottom-up manner. Common handcrafted features include center prior\cite{bib5} and distance transform\cite{bib6}, which only contain low-level clues and perform poorly in complex scenes. With the emergence of deep learning techniques, SOD has made remarkable progress. Recent SOD models are mainly implemented based on fully convolutional networks (FCNs) and Transformer architectures, among which FCN is still the mainstream SOD architecture. SOD FCNs-based models mainly design a series of decoders around feature extraction, refinement or enhancement, and multi-level feature fusion and set specific loss functions to assist model training\cite{bib7DC}\cite{bib8ICON}\cite{bib9DNTD}\cite{bib10EF}\cite{bib11MRI}\cite{bib12ED}\cite{bib13RCSB}\cite{bib14DP}\cite{bib15MSF}\cite{bib16PAKRN}. Inspired by the impressive performance of the Transformer in the field of natural language processing (NLP), researchers have also carried out some innovative work using the Transformer architecture in the SOD research field and achieved impressive performance improvements. These models mainly rely on the powerful long-distance feature correlation capture ability of the Transformer architecture and help the model generate more complete predictions by effectively extracting global contextual clues\cite{bib17VST}\cite{bib18LGVT}. In addition, a hybrid architecture of FCN and Transformer has been proposed by integrating the advantages of both models\cite{bib19SELF}\cite{bib20PG}.

The main research content of the above work is to design different network models carefully based on the characteristics of the SOD task and demonstrate the performance beyond the previous proposed models, playing an extremely important role in the rapid development process of the SOD research field. However, the process of carrying out these works is usually challenging, time-consuming, and requires difficultly obtaining higher performance than the latest methods. Therefore, in contrast to the above work, this paper proposes a positive feedback approach based on F-measure value for SOD, which is based on existing methods to obtain more accurate prediction results. The approach allows any existing method to be inserted, such as traditional algorithms and various deep learning methods, and can achieve performance beyond the inserted methods. 

In summary, the main contributions of this paper are:
\begin{itemize}
	\item Under the current SOD research background, we have explored how to utilize existing methods rarely. This paper proposes a positive feedback approach based on F-measure value for SOD, which allows any method to be inserted and performs better than the inserted method in terms of performance. The prediction process of the approach is training-free, adaptive weights, and contains only one tunable hyperparameter.
	\item Based on the F-measure value, a positive feedback process is designed, which does not require human involvement during the calculation process, and is a completely self-updating process.
\end{itemize}

\section{Related Work}\label{sec2}
Existing SOD models can be mainly divided into traditional algorithm-based and deep learning-based approaches, with the latter being the current mainstream SOD method.
\subsection{FCNs-based network}
FCNs-based SOD models mainly design a series of decoders around feature extraction, refinement or enhancement, and multi-level feature fusion and set specific loss functions to assist model training. 

The successful performance of SOD models depends on the effective integration of multi-level features. Several studies have highlighted the significant differences between the low-level and high-level features extracted from backbone networks. To address this issue, Chen et al.\cite{bib21GCPA} proposed a progressive context aware feature aggregation module, while Dai et al.\cite{bib10EF} developed a middle layer feature extraction module to achieve better feature fusion results. Moreover, the integration of multiscale features has been shown to improve SOD detection performance in various studies. For instance, Zhang et al.\cite{bib15MSF} developed a neural structure based search unit to automatically determine the multiscale features that need to be aggregated. Fang et al.\cite{bib9DNTD} proposed a densely nested-based network framework to utilize multiscale high-level feature maps effectively. Meanwhile, Zhuge et al.\cite{bib8ICON} used various multiscale feature extraction methods to obtain diverse multiscale features and designed an integrity channel enhancement module to highlight salient objects. Base on work\cite{bib22F3}, Wu et al. \cite{bib14DP} proposed a dynamic pyramid convolution to extract multiscale features instead of fixed size convolutions. Effective supervision strategy have also been employed to help models predict salient objects more accurately. For instance, Wei et al.\cite{bib22F3} proposed pixel position aware losss to guide networks to pay more attention to local details, while Yang et al.\cite{bib23PSGL} proposed progressive self-guided loss to guide network learning to more complete salient regions. Additionally, Xu et al.\cite{bib16PAKRN} designed a knowledge review network to roughly locate salient regions first and then finely segment salient objects. Wu et al. \cite{bib7DC} proposed a decomposition and completion network to predict the saliency, edge, and skeleton maps respectively, and then filled the saliency map using edge and skeleton maps.

\subsection{Transformer-based network}
In the field of SOD, the incorporation of global context information is extremely crucial for accurately and completely predicting salient regions. Transformer-based network architectures, renowned for their ability to capture long-range dependencies, have been successfully introduced in this domain.

To this end, Liu et al.\cite{bib17VST} proposed a unified transformer-based model for SOD, which facilitates the propagation of global context information among image patches. Meanwhile, Zhang et al.\cite{bib18LGVT} developed a generative vision transformer network that generates a pixel-level uncertainty map, effectively representing the significance confidence of salient objects. Additionally, Ren et al.\cite{bib24GLSTR} proposed a simple yet effective deeply-transformer network that preserves more unifying global-local representations to gradually restore spatial details.

\subsection{Hybrid framework-based network}
To enhance the accuracy and completeness of prediction results in the field of SOD, it is important to effectively extract and utilize both global context from deep features and local context from shallow features. However, achieving this in network frameworks can be challenging. Therefore, some hybrid network frameworks have emerged as the times require. 

Zhu et al.\cite{bib25DFTR} proposed a deep supervised fusion transformer network, extending the applicability of FCN to the transformer architecture for the first time. They employed a transformer encoder to extract multiscale features and designed a multiscale aggregation module to aggregate these features in a coarse-to-fine manner. Similarly, Yun et al.\cite{bib19SELF} proposed a self-refined transformer network that leverages the transformer encoder to capture long-distance dependencies and designed a context refinement module. The module is employed to integrate global context with decoder features and refine and locate local details automatically. In contrast, Wang et al.\cite{bib26TF} still used FCN as an encoder and utilized a transformer module for multi-level feature fusion to address the limited receptive field of FCN.


\section{Method}\label{sec3}
\subsection{Overview of the proposed Method}
\begin{figure*}[!htb]
	\centering
	\includegraphics[scale=0.6]{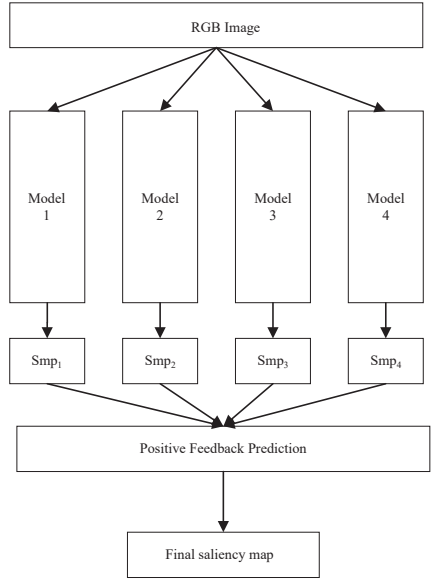}	
	\caption{\label{fig-pf}The overall pipeline of the proposed method. }
\end{figure*}

Figure \ref{fig-pf} shows structure of the proposed method, which includes a multi-branch model structure and a positive feedback prediction structure. The former is used to place existing SOD model sequences, including traditional algorithm models, deep-learning supervised or unsupervised models. The positive feedback prediction structure will iteratively calculate the respective weights according to the outputs of all branches to obtain the final saliency maps, where the weight calculation process is completely adaptive. In order to verify our method more conveniently, we use the precious models in advance to generate saliency maps, which are marked as $Smp_i(i=1,2,3,4)$.

%
\subsection{Positive feedback prediction algorithm}
\begin{figure*}[h]
	\centering
	\includegraphics[width=1\textwidth]{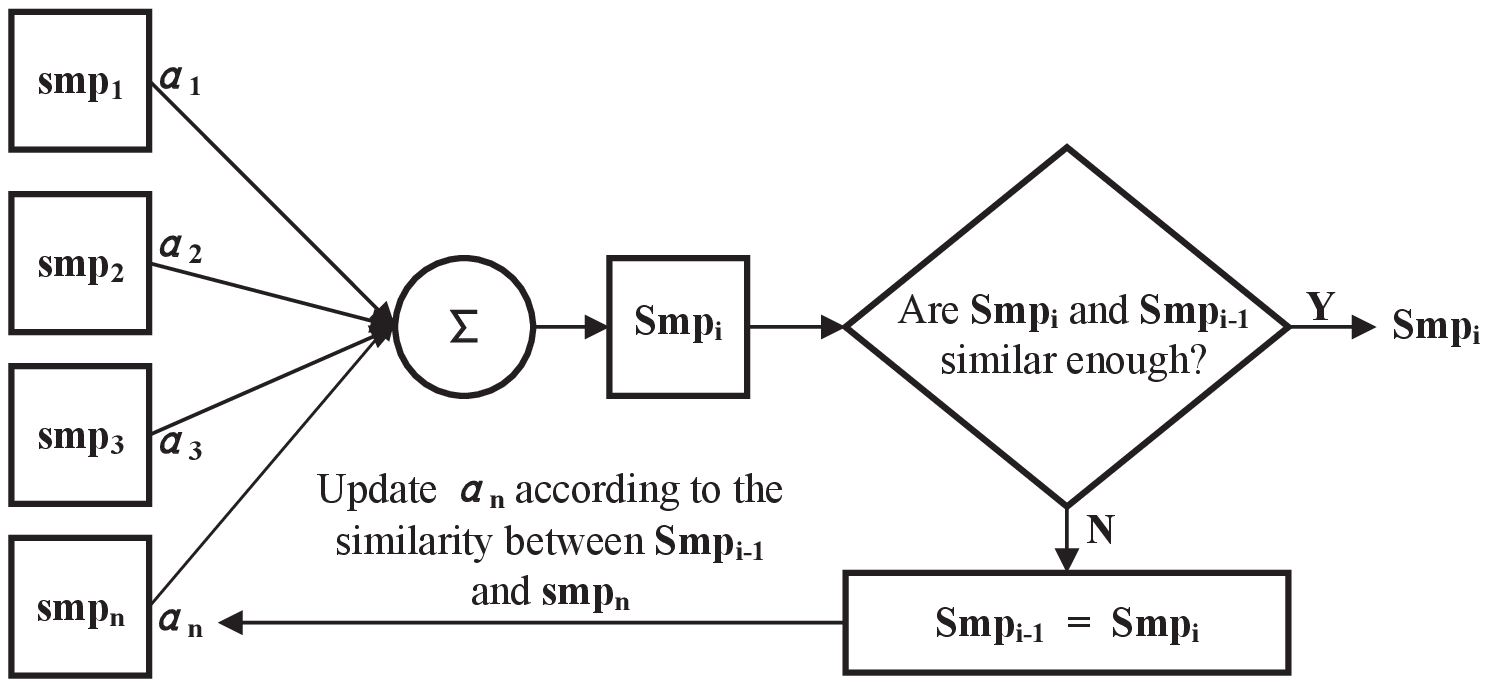}
	\caption{The conceptual principle of positive feedback prediction algorithm.}
	\label{Algorithm1-fig}
\end{figure*}

The positive feedback prediction algorithm is summarized in Table \ref{Algorithm1-tab}, and its conceptual principle is illustrated in Figure \ref{Algorithm1-fig}.  Hereinafter, $smp_n, N$ refers to the saliency maps generated by the multi-branch models and the number of branch. $\varepsilon$ is a  threshold, which is set to 0.95. The $imbinarize$($\cdot$) is a binarization function and the $mat2grey$ is a graying function. The $computerFmeasure$($\cdot$) is the function for calculating $F$-$measure$\cite{bib27}. 

The key steps of the algorithm: 
\newline \textbf{Step1.} Input: outputs ($smp_n$) of the multi-branch model structure are used as inputs of the positive feedback prediction structure, and perform the binarization function to get binary maps ($b\_smp_n$).
\newline \textbf{Step2.} Initialization: the above inputs are fused in the way of pixel-level addition, and perform the binarization function to get binary maps ($B\_Smp_0$). At this time, the weights of each branch are consistent.
\newline \textbf{Step3.} Iteration: calculate $F$-$measure$ with the input ($b\_smp_n$) of each brach and the latest fusion result ($B\_Smp_{i-1}$). Update the weights of each branch, recalculate the fusion result ($Smp_{i}$) from the new weights  and perform the binarization function to get binary maps ($B\_Smp_{i}$).
\newline \textbf{Step4.} Judgement: calculate $F$-$measure$ with the latest and last fusion result and compare it with $\varepsilon$. If it is greater than the threshold, the two successive resluts are considered to be simmilar enough, and the latest fusion result is taken as the final output; Otherwise, execute \textbf{Step3} again.

\begin{table*}[ht]		
	\caption{Positive feedback prediction algorithm.}
	\label{Algorithm1-tab}
	\begin{tabular}{l l}			
	    \hline
		\textbf{Input}:  	&       \\
		&	$smp_n, N$;  \\
		\textbf{Output}: 	&       \\
		&   $Smp_{i}$; \\
		1:      &  $B\_Smp_{i-1} = imbinarize(mat2grey(\sum_{n=1}^{N}smp_n)), i = 1 $; \\
		2:      &  $b\_smp_n = imbinarize(smp_n) $;\\
		3:      &  \textbf{while} True \textbf{do} \\
		4:      &  \quad\quad $F_n =computerFmeasure(b\_smp_n, B\_Smp_{i-1})$; \\
		5:      &  \quad\quad $F_{sum} = \sum_{n=1}^{N}F_n$; $\alpha_n = {F_n}\setminus{F_{sum}}$; \\
		6:      &  \quad\quad $Smp_i = \sum_{n=1}^{N}{\alpha_n} \cdot smp_n;B\_Smp_i = imbinarize(Smp_i)$; \\
		7:      &  \quad\quad $F =computerFmeasure(B\_Smp_i, B\_Smp_{i-1})$; \\
		8:      &  \quad\quad  \textbf{if} $F \geq \varepsilon$ \textbf{then} \\
		9:      &  \quad\quad\quad\quad $break$; \\
		10:      &  \quad\quad  \textbf{else} \\
		11:      &  \quad\quad\quad\quad $i = i + 1$; \\
		12:      &  \quad\quad \textbf{end if} \\
		13:      &  \textbf{end while} \\
		14:      &  $Smp_{i} = mat2grey(Smp_{i})$; \\
		15:      &  \textbf{return}\quad $Smp_{i}$. \\		
		\hline
	\end{tabular}
\end{table*}

\section{Experimental results and analysis}\label{sec4} %
\subsection{Datasets and Evaluation Metrics}
To validate the efficacy of the proposed method in this paper, we performed a series of experiments on five publicly available datasets. The datasets utilized in this study are briefly described as follows: DUTS\cite{bib28} comprises a total of 10,000 images for training and 5,000 images for testing, with only the test set being used in this study. DUT-OMRON\cite{bib29} consists of 5,019 images with complex structures and backgrounds. HKU-IS\cite{bib30} contains 4,447 maps with multiple salient objects. PASCAL-S\cite{bib31} contains 850 natural images. ECSSD\cite{bib32} comprises a collection of 1,000 images obtained from the internet.

To quantitatively evaluate the proposed method, we adopt six evaluation metrics as the performance measures, including mean absolute error ($MAE$)\cite{bib33}, maximum F-measure ($mF$) score, and S-measure ($Sm$) score\cite{bib34}, precision-recall ($PR$) curves.

\subsection{Implementation Details}
To facilitate the testing of the proposed positive feedback method, we implemented the following steps. Firstly, we generated prediction maps of a multi-branch model using the model disclosed by the author and Python tools. Secondly, we evaluated the positive feedback method using Matlab tools.

To evaluate the effectiveness and characteristics of our proposed method, we conducted comparative, ablation, and robustness experiments. In the comparative experiment, we compared the performance with other state-of-the arts models. Results showed that the positive feedback mechanism improved the model's performance on multiple metrics. In the ablation experiment, we compared the performance of the positive feedback mechanism with pixel-level addition fusion on five datasets. In the robustness experiment, we manually selected good and bad renderings from each model and input them into the positive feedback mechanism to observe the visual comparison results.

In total, we conducted two sets of experiments. In the first set, we used four branches, namely MSFNet (2021)\cite{bib15MSF}, PAKRNet (2021)\cite{bib16PAKRN}, ICONet (2022)\cite{bib8ICON}, and SelReformer (2022)\cite{bib19SELF}. The performance of the 2021 methods are weaker than that of the 2022 methods. In the second set, we used two branches, namely DPNet (2022)\cite{bib14DP} and SelReformer. The performance of these two methods is relatively close. For detailed experimental procedures and analysis, please refer to each subsection.  

\begin{table*}[h]
	\fontsize{11}{13.2}\selectfont
	\caption{Quantitative evaluation. The mean absolute error (MAE, smaller is better), maximum F-measure (mF, larger is better) and S-measure (Sm, larger is better). The results for each saliency detection method are reported on five different datasets, with the top three performing methods highlighted in red, green, and blue. It is important to note that "Our-SSSS" and "Our-SS" denote the results of the four-branch and two-branch models, respectively.}
	\label{comparison-metrics}
	\centering
	\resizebox{1\textwidth}{!}	{
	\begin{tabular}{l c@{ }c@{ }c c@{ }c@{ }c c@{ }c@{ }c c@{ }c@{ }c c@{ }c@{ }c}
		\toprule
		\multirow{2}{*}{Methods} &\multicolumn{3}{c}{DUTS} &\multicolumn{3}{c}{ECSSD} &\multicolumn{3}{c}{HKU-IS}& \multicolumn{3}{c}{PASCAL-S} & \multicolumn{3}{c}{DUT-OMRON}\\
		
		\cline{2-16} 				& mF &MAE &Sm   	& mF &MAE  &Sm   	& mF &MAE  &Sm   	& mF &MAE  &Sm	 	& mF &MAE  &Sm \\
		\midrule
		
		DCNet\cite{bib7DC}          &.894&.035&.895		&.952&.032&.928	    &.939&.027&.922 	&.872&.062&.861  	&.823&.051&.845\\
		BiconNet\cite{bib35}        &.888&.038&.890 	&.949&.034&.927 	&.939&.029&.923 	&.877&.063&.863 	&.817&.053&.842\\
		DNTD\cite{bib9DNTD}         &.892&.033&.891 	&.946&.034&.922 	&.938&.028&.920 	&.878&.064&.857  	&.803&.051&.828\\
		EDNet\cite{bib12ED}         &.895&.035&.892 	&.951&.032&.927 	&.941&.026&.924		&.886&.062&.865  	&.828&.049&.850\\
		EFNet\cite{bib10EF}         &.898&.034&.895 	&.948&.034&.925 	&.939&.027&.922 	&.875&.063&.864  	&.822&.054&.843\\
		MRINet\cite{bib11MRI}       &.899&.035&.894 	&.950&.032&.927 	&.941&.027&.922		&.877&.060&.864 	&.829&.054&.848\\
		RCSBNet\cite{bib13RCSB}     &.899&.035&.881 	&.944&.034&.922 	&.938&.027&.919 	&.882&.059&.860  	&.810&.049&.835\\
		\hline
		MSFNet\cite{bib15MSF}       &.878&.034&.877 	&.941&.033&.915		&.927&.027&.908	 	&.863&.061&.852 	&.799&.050&.832\\
		PAKRNet\cite{bib16PAKRN}    &.907&.033&.900 	&.953&.032&.928		&.943&.027&.924		&.873&.067&.858		&.834&.050&.853\\			ICONet\cite{bib8ICON}       &.893&.037&.892 	&.951&.031&.931		&.942&.027&.925	 	&.884&.060&.870		&.830&.059&.846\\
		
		DPNet\cite{bib14DP}         &\textcolor{blue}{.917}&\textcolor{green}{.028}&\textcolor{green}{.912} 	&.954&.031&.931 	&\textcolor{green}{.950}&\textcolor{red}{.023}&\textcolor{blue}{.934} 	&\textcolor{blue}{.894}&\textcolor{blue}{.054}&.877  &.834&.049&.853\\

		SelReformer\cite{bib19SELF} &.916&\textcolor{red}{.027}&\textcolor{blue}{.911} 	&\textcolor{blue}{.958}&\textcolor{red}{.027}&\textcolor{blue}{.936} 	
		&\textcolor{blue}{.947}&\textcolor{green}{.024} &.931 
		&\textcolor{blue}{.894}&\textcolor{red}{.051}&\textcolor{blue}{.881}  	
		&\textcolor{blue}{.837}&\textcolor{red}{.043}&\textcolor{blue}{.861}\\
			
		Our-SSSS   					&\textcolor{green}{.922}&\textcolor{blue}{.032}&.909 				&\textcolor{green}{.960}&\textcolor{blue}{.029}&\textcolor{green}{.941} &\textcolor{green}{.950}&\textcolor{blue}{.025}&\textcolor{green}{.935} 	
		&\textcolor{green}{.898}&.055&\textcolor{green}{.884}  	
		&\textcolor{red}{.863}&\textcolor{red}{.043}&\textcolor{red}{.877}\\
		
		Our-SS          			&\textcolor{red}{.931}&\textcolor{green}{.028}&\textcolor{red}{.919}	&\textcolor{red}{.962}&\textcolor{green}{.028}&\textcolor{red}{.943} 
		&\textcolor{red}{.954}&\textcolor{red}{.023}&\textcolor{red}{.939} 	
		&\textcolor{red}{.899}&\textcolor{green}{.052}&\textcolor{red}{.889} &\textcolor{green}{.859}&\textcolor{green}{.046}&\textcolor{green}{.869}\\					
		\bottomrule
	\end{tabular}
	}
\end{table*}

\subsection{Comparison with the State-of-the-arts} 
We compare the proposed method with 12 state-of-the-art methods, including DCNet\cite{bib7DC}, BiconNet\cite{bib35}, DNTD\cite{bib9DNTD}, EDNet\cite{bib12ED}, EFNet\cite{bib10EF}, MRINet\cite{bib11MRI}, RCSBNet\cite{bib13RCSB}, MSFNet\cite{bib15MSF}, PAKRNet\cite{bib16PAKRN}, ICONet\cite{bib8ICON},
DPNet\cite{bib14DP}, SelReformer\cite{bib19SELF}. To ensure a fair and objective comparison, we will utilize the saliency maps provided by the authors and employ the same standardized evaluation function to calculate each metric.

\begin{figure*}[h]
	\centering
	\includegraphics[width=1\textwidth]{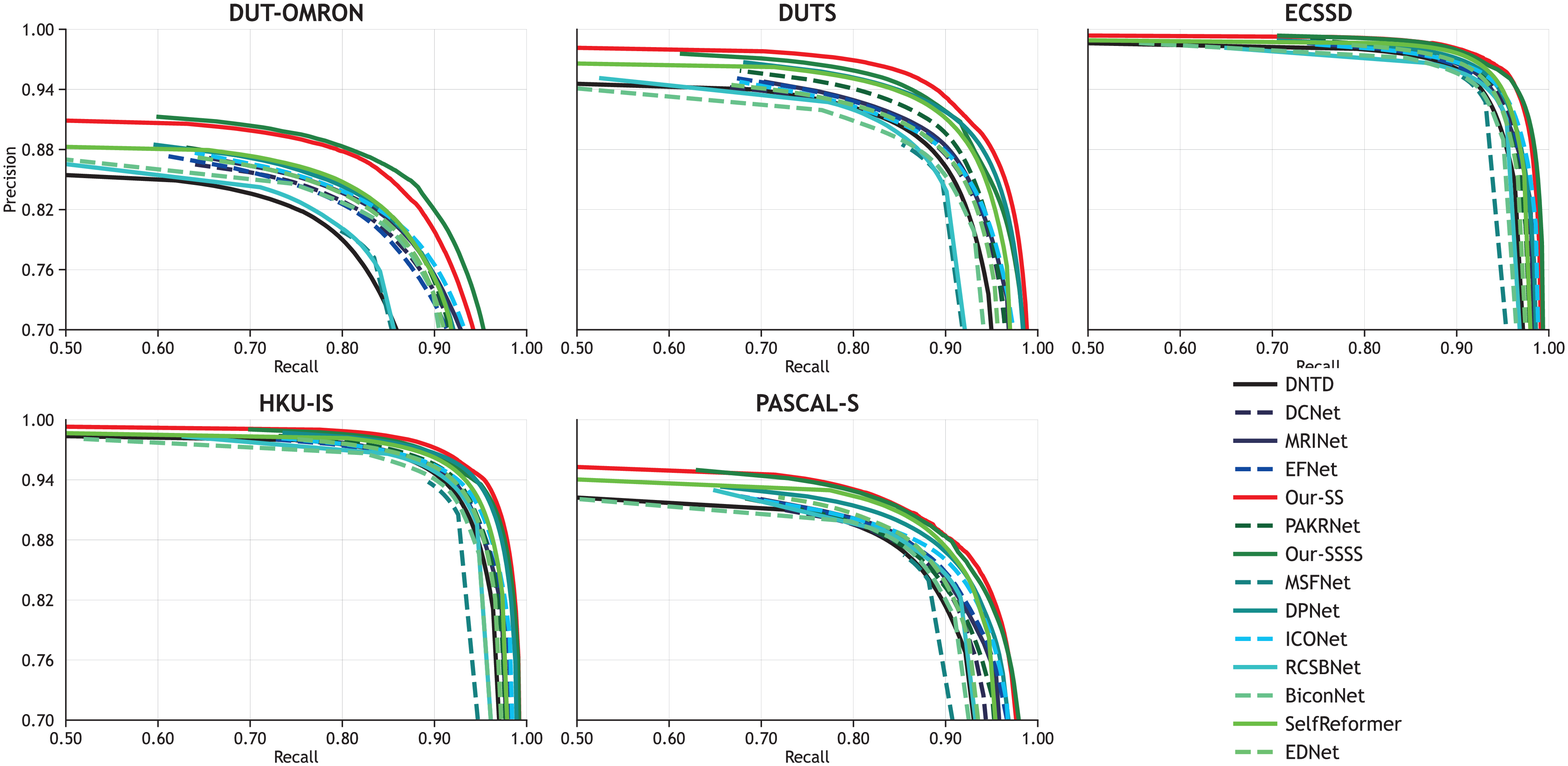}
	\caption{Performance comparison with 15 state-of-the-art methods on five saliency datasets. The red solid line and the blue solid line denote our models.}
	\label{fig-pr}
\end{figure*}

In Table \ref{comparison-metrics}, we present the quantitative comparison results based on MAE, maximum F-measure, and S-measure. Our methods demonstrate the most comprehensive performance across all five datasets, as evidenced by our superior results across all three metrics. Notably, our method, Our-SS, consistently outperforms the second-best result in terms of mF and Sm by 1.5\%, 0.4\%, 0.4\%, 0.5\%, and 0.9\%, 0.8\%, 0.6\%, 0.9\% on DUTS, ECSSD, HKU-IS, and PASCAL-S datasets, respectively. Additionally, our method, Our-SSSS, performs exceptionally well on ECSSD, HKU-IS, and PASCAL-S datasets, particularly on the DUT-OMRON dataset, where it consistently outperforms SelReformer\cite{bib19SELF}, the strongest model among the four-branch structures. Moreover, in Figure \ref{fig-pr}, we provide precision-recall curves (PR) for all five datasets. Our curves demonstrate exceptional performance across most thresholds, particularly on the DUT-OMRON and DUTS datasets.

\subsection{Ablation experiment}
\indent\indent Within this subsection, we have performed ablation experiments on two datasets to demonstrate the efficacy and distinctive attributes of the proposed approach. Specifically, we have replaced the proposed positive feedback fusion with pixel-level addition and have presented the experimental outcomes in Table \ref{tab-ablation}. The results reveal that our proposed method yields superior overall performance compared to the direct addition way. Notably, in the ablation experiment pertaining to the four-branch architecture, our proposed approach has demonstrated a considerably more substantial advantage, especially for the mean absolute error (MAE) metric, which has reduced by 7.2\%and 7.7\%, correspondingly.

\begin{table}[h]
	\fontsize{11}{13.2}\selectfont
	\caption{Ablation evaluation on two datasets.}	
	\label{tab-ablation}
	\centering
	\resizebox{0.5\textwidth}{!}{
		\begin{tabular}{l c@{ }c@{ }c c@{ }c@{ }c}
			\hline
			\multirow{2}{*}{Methods} &\multicolumn{3}{c}{PASCAL-S} &\multicolumn{3}{c}{DUT-OMRON}\\		
			\cline{2-7} 			& mF &MAE  &Sm   	& mF &MAE  &Sm   	\\
			\hline			
			ablation-4S     		&.891&.060&.882 	&.863&.047&.874 	\\
			Our-SSSS     			&.898&.055&.884 	&.863&.043&.877 	\\
			\hline 
			\hline
			ablation-2S   			&.897&.053&.888 	&.858&.046&.869 	\\	
			Our-SS          		&.899&.052&.889		&.859&.046&.869 	\\		
			\hline \\			
		\end{tabular}
	}
\end{table}

Based on the results of the two sets of comparative experiments on ablation, we can derive several professional conclusions. Firstly, in cases where there is a significant difference in performance between the selected methods and their placement in a multi-branch structure, the proposed method exhibits greater advantages over direct fusion. Conversely, when the performance difference between the selected methods is minor, the proposed method can only exhibit limited superiority over direct fusion. Additionally, it is important to note that the performance synthesis level achieved through positive feedback fusion is consistently superior to that of each individual model within a multi-branch structure. 

\subsection{Robustness experiment}
\begin{figure*}[h]
	\centering
	\includegraphics[width=1\textwidth]{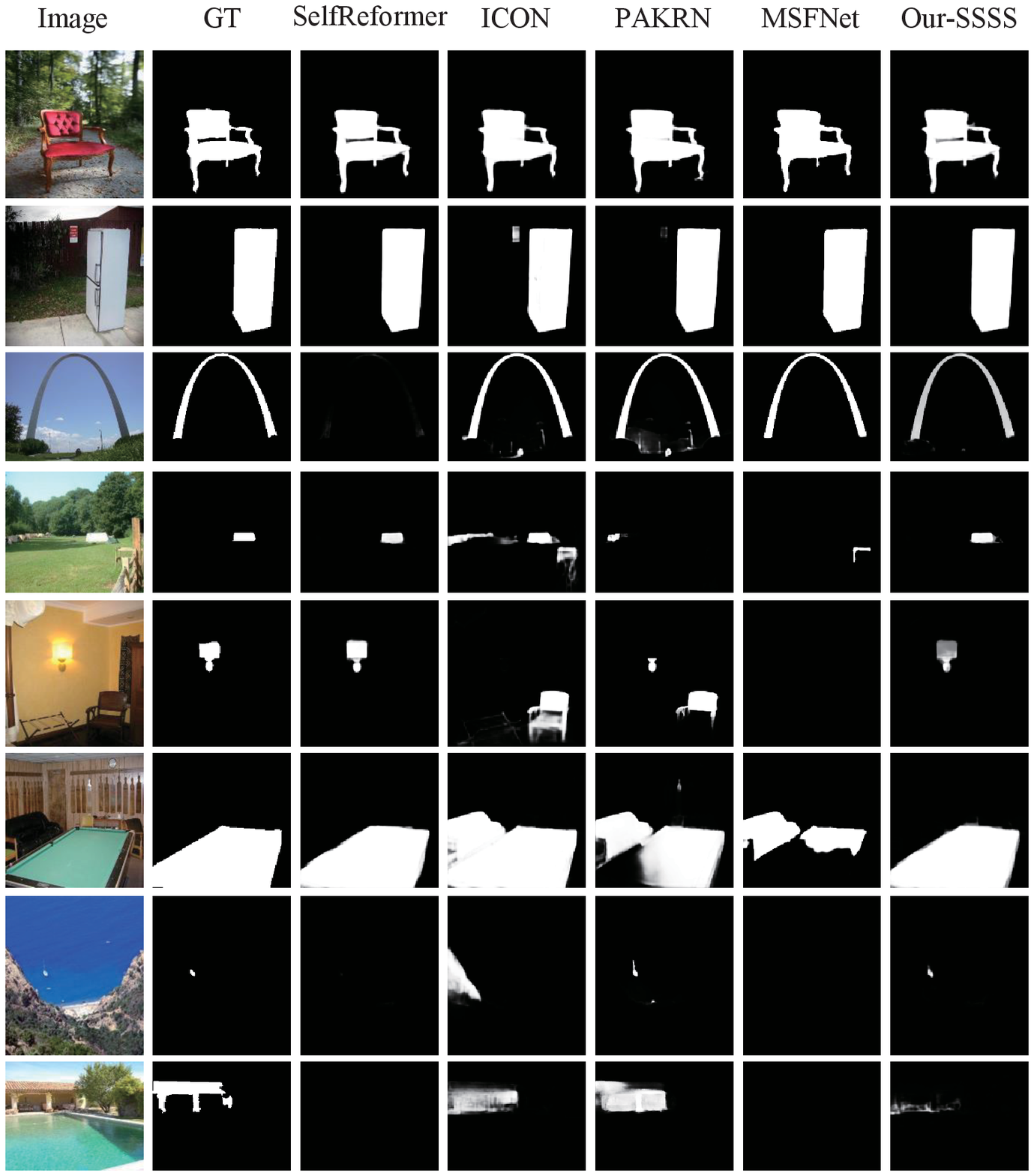}
	\caption{Visual comparisons from the four-branch structure.}
	\label{fig-vision}
\end{figure*}

To assess the proposed method's robustness against interference, we conducted adversarial experiments on a four-branch structure. Despite the outstanding detection performance of many existing methods, their capabilities are not consistently strong, as depicted in Figure \ref{fig-vision}. To better showcase the superiority of our approach, we manually selected several sets of input predictions with both desirable and undesirable effects for the proposed method, and presented visual comparisons in Figure \ref{fig-vision}. This not only explains why the proposed method is superior to methods introduced into a multi-branch structure, but also superior to direct fusion, as it allows for automatic weight updating to achieve satisfactory results, instead of treating all branches equally. Furthermore, it should be noted that even in the absence of good predictions in the four-branch structural model, a favorable outcome may still be yielded, as demonstrated in the seventh row.

\section*{Conclusion}\label{sec5}
Distinguished from most existing efforts in salient object detection (SOD), this study focuses on enhancing the accuracy of previously established algorithms. Specifically, we introduce a positive feedback approach based F-measure value for SOD that comprises a multi-branch model structure and a positive feedback prediction structure. The method involves feeding input images into the multi-branch model structure to generate their corresponding saliency maps, and then processing these maps through the positive feedback prediction structure for obtaining the final result via positive feedback calculation. Notably, our method requires no model training, entails minimal hyperparameter tuning, and features automatic calculation. By integrating multiple existing models, our method surpasses the performance of individual models. Our proposed method achieves a prediction speed of 20 frames per second (FPS) even on a low-end host, while removing the prediction time overhead of inserted models. Our findings suggest that higher performance of the inserted models leads to better overall results, and the initialization stage is crucial due to the positive feedback mechanism. Selecting models with insignificant performance differences weakly showcases the advantages of our approach. We believe that our study is not only of significant research value but also has practical applications, and thus, merits the attention and scrutiny of future researchers.



\section*{Conflict of interests}

The authors declare that they have no conflict of interests.

\section*{Acknowledgments}
This work was supported by Key Research and Development Projects in Zhejiang Province of China (NO.2021C03192, 2023C01032).

\section*{Author contributions}
Ailing Pan wrote the paper, conceived and designed the experiments; \\
Chao Dai assisted in revising paper, collected the data, performed the experiments and recorded experimental results; \\
Chen Pan proposed the core idea of the paper and will act as the corresponding author; \\
Dongping Zhang and Yunchao Xu are participants in the fund projects. \\
All authors agree to this submission.

\section*{Code and data availability}
Code and Saliency maps will be available at: https://github.com/dc3234/PF/tree/main.

\end{document}